\documentclass[journal]{IEEEtran}
\ifCLASSINFOpdf
\else
\fi
\usepackage{url}
\usepackage{amsmath,graphicx}
\usepackage{amsfonts}
\usepackage{amssymb}
\usepackage{booktabs}
\usepackage{multirow}
\usepackage{bm}
\usepackage[linesnumbered, lined, ruled]{algorithm2e}

\begin{document}
%
\title{BiCANet: Bi-directional Contextual Aggregating Network for Image Semantic Segmentation}
%
%
%

\author{Quan Zhou,~\IEEEmembership{Member,~IEEE,}
        Dechun Cong, Bin Kang, Xiaofu Wu, Baoyu Zheng,~\IEEEmembership{Seniour Member,~IEEE,} \\ 
        Huimin Lu,~\IEEEmembership{Seniour Member,~IEEE,} and Longin Jan Latecki,~\IEEEmembership{Seniour Member,~IEEE}
\thanks{Manuscript received XXXX XX, 2020; revised XXXX XX, 2020; accepted XXXX XX, 2020. This work was jointly supported in part by the National Natural Science Foundation of China under Grants 61876093, 61671253, the National Natural Science Foundation of Jiangsu Province under Grant BK20181393, the National Science Foundation under Grant IIS-1302164, and in the part by the China Scholarship Council under Grant 201908320072.}
\thanks{\emph{Corresponding author: Quan Zhou.}}
\thanks{Quan Zhou, Dechun Cong, Xiaofu Wu, and Baoyu Zheng are with National Engineering Research Center of Communications and Networking, Nanjing University of Posts \& Telecommunications, Nanjing, Jiangsu, P. R. China. (e-mail: quan.zhou@njupt.edu.cn, 1121781207@qq.com, xfuwu@njupt.edu.cn, zby@njupt.edu.cn)}
\thanks{Bin Kang is with the Department of Internet of Things, Nanjing University of Posts \& Telecommunications, Nanjing, Jiangsu, P. R. China. (e-mail: kangb@njupt.edu.cn)}
\thanks{Huimin Lu is with the Department of Mechanical and Control Engineering, Kyushu Institute of Technology, 	Kitakyushu, Fukuoka, Japan. (e-mail: dr.huimin.lu@ieee.org)}
\thanks{Longin Jan Latecki is with the Department of Computer and Information Sciences, Temple University, Philadelphia, Pennsylvania, USA. (e-mail: latecki@temple.edu)}
}

%
%

\markboth{IEEE Transactions on Image Processing,~Vol.~XX, No.~XX, January~2020}%
{Quan \MakeLowercase{\textit{et al.}}: BiCANet: Bi-directional Contextual Aggregating Network for Image Semantic Segmentation}
%



\maketitle

\begin{abstract}

Exploring contextual information in convolutional neural networks (CNNs) has gained substantial attention in recent years for semantic segmentation. This paper introduces a Bi-directional Contextual Aggregating Network, called \emph{BiCANet}, for semantic segmentation. Unlike previous approaches that encode context in feature space, \emph{BiCANet} aggregates contextual cues from \emph{a categorical perspective}, which mainly consists of three parts: contextual condensed projection block (CCPB), bi-directional context interaction block (BCIB), and muti-scale contextual fusion block (MCFB). More specifically, CCPB learns a category-based mapping through a split-transform-merge architecture, which condenses contextual cues with different receptive fields from intermediate layer. BCIB, on the other hand, employs dense skipped-connections to enhance the class-level context exchanging. Finally, MCFB integrates multi-scale contextual cues by investigating short- and long-ranged spatial dependencies. To evaluate \emph{BiCANet}, we have conducted extensive experiments on three semantic segmentation datasets: PASCAL VOC 2012, Cityscapes, and ADE20K. The experimental results demonstrate that \emph{BiCANet} outperforms recent state-of-the-art networks without any postprocess techniques. Particularly, \emph{BiCANet} achieves the mIoU score of  86.7\%, 82.4\% and 38.66\% on PASCAL VOC 2012, Cityscapes and ADE20K testset, respectively. Our code is open-source and is publicly available at \url{https://github.com/cdcnjupt/BCANet}.

\end{abstract}


%
\IEEEpeerreviewmaketitle

\section{Introduction}\label{sec:Introduction}
%
%
%
%
\IEEEPARstart{S}{emantic} segmentation plays an important role in image understanding, and facilitates many real-world applications such as self-driving, human-machine interaction, and retrieval-based searching engines. The goal of semantic segmentation aims at assigning a categorical label to each image pixel, which thus can be also considered as a dense prediction problem. From the perspective of computer vision, there are two major sub-tasks for image semantic segmentation: (1) classification, where a unique semantic concept should be marked correctly to the associated object; (2) localization, where the assigned label for pixel must be aligned to the appropriate coordinates in the segmentation output. To this end, a well-designed segmentation system should simultaneously deal with these two issues.

Due to the powerful ability to abstract high-level semantics from raw images, the recent years have witnessed remarkable progress for the task of semantic segmentation using deep convolutional neural networks (CNNs). As a pioneer work, Long \emph{et al.} \cite{long2017fully} proposed a VGG-based fully convolutional neural networks (FCNs), which adopts filtering convolution for 2D predictions instead of fully connected network used for classification. After that, a vast number of FCNs has been proposed for semantic segmentation \cite{chao2017large,zhao2017pyramid,Chen2016deeplab,Badrinarayanan2015Segnet,fu2019dual}. However, multiple stages of spatial pooling and convolution stride significantly reduce the dimension of feature representation, thereby losing much of the finer image structure. This invariance to local image transformation is helpful for image classification \cite{he2016res,going2015szegedy}, but may be harmful for dense estimation problems, especially the task of semantic segmentation \cite{chao2017large,Chen2016deeplab}. An alternative approaches to address this problem are using encoder-decoder networks (EDNs), where the high-resolution feature maps are sequentially recovered by learning deconvolutional filters \cite{Badrinarayanan2015Segnet,noh2015learning,Guosheng2017RefineNet}. These approaches, however, still suffer from a couple of critical limitations. Firstly, due to the significant loss of spatial information in the encoder stage, the deconvolutions are often hard to recover the details of low-level visual features, leading to the degradation of performance \cite{chao2017large,Guosheng2017RefineNet}. Secondly, training EDNs is a nontrivial work since they are nearly twice deeper than FCNs \cite{Badrinarayanan2015Segnet,noh2015learning}.

In order to overcome these challenges, the recent research focuses on designing network architecture to capture context information for semantic segmentation \cite{long2017fully,zhao2017pyramid,Chen2016deeplab,feed2015mos, exploring2018lin,zhang2019co,ding2019se}. For example, the variants of FCNs \cite{long2017fully,Badrinarayanan2015Segnet,bilin2018dense} employ skipped-connections to encode context clues for high-resolution estimation, where the intermediate layer features are investigated. In contrast, the dilated/atrous convolution \cite{Chen2016deeplab,yang2018dense,multi2016yu} carefully designs convolutional filters to enlarge receptive field, enabling the CNNs sensitive to global context semantics. However, solely depending on very few surrounding spatial positions may exhibit limited representation power of contextual features, thus always leading to the gridding artifacts in filtering responses. Additionally, as a widely-used formulation that leverages context information in graph modeling, Conditional Random Fields (CRFs) meet their opportunities to integrate with CNNs for semantic segmentation \cite{exploring2018lin,vem2016gaussian,conditional2015zheng,liu2018deep}, where mean field inference is treated as recurrent layers in an end-to-end training manner. An alternative approaches to aggregate context are employing larger convolutional kernels \cite{chao2017large,zhao2017pyramid,he2019ada}. Using a single convolutional layer, however, may be inadequate to cover correlated areas, resulting in the problem that objects substantially larger or smaller than the receptive field may be fragmented or incorrectly classified \cite{Chen2016deeplab,Badrinarayanan2015Segnet,noh2015learning}. Most recently, visual attention has gained substantial focus to capture long-ranged dependencies for semantic segmentation \cite{fu2019dual,li2018pyr,zhao2018psanet}, where soft-attention \cite{hu2018se} utilizes the pooling operation to squeeze global context used for reweighting channel-wise attention \cite{li2018pyr,zhang2018con} and self-attention \cite{vas2017att} harvests various range contextual information by producing a pixel-wise attention \cite{zhao2018psanet,vas2017att}. In spite of achieving promising results, these methods inherently suffer from following limitations. Firstly, the previous CNN-based approaches learn a powerful context representation in the feature space, lacking the consideration of \emph{explicitly} capturing contextual clues from a \emph{categorical perspective}. Secondly, the proposed CNNs aggregate context in a non-adaptive manner, neglecting the fact that different pixels require different ranged dependencies. Finally, although there are many successful CNNs proposed for context aggregation, it is still hard to design CNN architecture that integrates different ranged dependencies in an appropriate way.

This paper designs a Bi-directional Contextual Aggregating Network, called \emph{BiCANet}, which makes an effort to address above challenges. In contrast to previous literature \cite{Chen2016deeplab,zhang2019co,ding2019se,shi2018hier,ding2020se} that prefers to \emph{implicitly} learn powerful context representation in feature space, our method \emph{explicitly} explores category-based context, which is often ignored yet plays an essential role in semantic segmentation. Specifically, our \emph{BiCANet} mainly consists of three components: contextual condensed projection block (CCPB), bi-directional context interaction block (BCIB), and multi-scale contextual fusion block (MCFB). At the beginning, from each intermediate convolution stage, CCPB learns a category-based projection through a multiple branch architecture, where a split-transform-merge scheme is utilized to condense contextual cues with adaptive receptive fields. In order to enhance the information exchange among different intermediate layers, BCIB designs a context interaction subnetwork using dense bi-directional skipped-connections, harvesting fine details and class-level semantics that complement each other to boost performance. Finally, solely relying on fixed scale context is not enough for dense estimation problems, correctly identifying image pixels needs multi-scale context information from local surroundings, long-ranged dependencies, even the entire scene. To this end, we put forward MCFB to fulfill this task where appropriate contextual cues are adaptively selected for different pixels.

One main merit of \emph{BiCANet} is the fact that it allows us to explicitly integrate category-based estimations between pixels into an optimization problem, which can be considered as simulating graphical inference (e.g., CRF) into a single unified compact model. Moreover, this approach can be also viewed as an appropriate way, which explicitly learns an adaptive and robust context representation to make prediction for each pixel, not only based on local information but also on long-ranged surroundings, and even global context from the entire scene.

Another advantage of our method lies in the fact that it is very simple, flexible, and effective. There are no postprocess techniques involved in our approach, yet it does not exclude recent advances, such as image pyramid \cite{Guosheng2017RefineNet}, CRFs \cite{Chen2016deeplab,exploring2018lin,liu2018deep}, and additional segmenting losses \cite{zhao2017pyramid}, to further improve performance. Additionally, our approach can be easily plugged into advanced feature abstraction backbones, such as VGGNet \cite{long2017fully}, ResNet \cite{he2016res}, ResNext \cite{xie2017agg}, and DenseNet \cite{huang2017den}, without introducing additional operations. We have evaluated our \emph{BiCANet} on three semantic segmentation datasets: PASCAL VOC 2012 \cite{mark2015pascal}, Cityscapes \cite{Cordts2016the}, and ADE20K \cite{zhou2017sce}, and the experimental results show the superior performance of our method with respect to recent state-of-the-art networks. In summary, the main contributions of this paper are four-fold:

\begin{itemize}
	\item {We propose a novel CCPB, allowing us to \emph{explicitly} investigate context from \emph{categorical perspective}, which has been often ignored yet plays a significant role for the task of semantic segmentation.}
	\item {Based on CCPB, we put forward a context exchanging network, named BCIB, through bi-directional skipped-connections,  making full use of contextual cues from intermediate convolution layers.}
	\item {In order to assign appropriate contextual clues to different pixels, MCFB is adopted in our \emph{BiCANet} to \emph{explicitly} explore multi-scale context by encoding short-ranged, long-ranged, and global dependencies.}
	\item {We test \emph{BiCANet} on three benchmarks. The comprehensive experiments demonstrate that our approach achieves state-of-the-art results. Particularly, \emph{BiCANet} achieves 86.7\%, 82.4\% and 38.66\% mIoU for PASCAL VOC 2012 \cite{mark2015pascal}, Cityscapes \cite{Cordts2016the}, and ADE20K \cite{zhou2017sce}, respectively.}
\end{itemize}

An early version of this work was first published in \cite{cong2019can}. This journal version extends previous one in following aspects: (1) Instead of using VGG-16 \cite{long2017fully}, we employ ResNet \cite{he2016res} as more powerful feature abstracting backbone; (2) An enhanced version of contextual condensed block is proposed, in which a split-transform-merge scheme is adopted to adaptively enlarge receptive field, enabling \emph{BiCANet} to capture context information from categorical perspective; (3) Unlike \cite{cong2019can} that directly concatenates condensed outputs, \emph{BiCANet} designs a dense bi-directional interaction network that exchanges context information from different intermediate layers; (4) The previous version directly transfers stacked features to softmax classifier. In contrast, our \emph{BiCANet} utilizes MCFB to explicitly capture context from local to global dependencies, arranging adaptive contextual dependencies for different pixels; (5) We have performed more exhausted evaluations and ablation experiments, and report more comparisons and improved results.

The remainder of this paper is organized as follows. After a brief introduction of related work in Section \ref{sec:Relatedwork}, we elaborate on the details of our \emph{BiCANet} in Section \ref{sec:BiCANet}. Experimental results are given in Section \ref{sec:Experimence}, and Section \ref{sec:Conclusion} provides conclusion remarks and future work.

\section{Related Work}\label{sec:Relatedwork}

In recent literature, there are vast number of deep neural networks that capture context information for semantic segmentation  \cite{long2017fully,Chen2016deeplab,zhang2019co,ding2019se,yang2018dense,ding2020se,zhou2019con}. Immediately below, we review the related works, which can be roughly divided into three categories: EDNs, context aggregation networks (CANs), and attention embedding networks (AENs).

\subsection{EDNs}

EDNs often utilize skipped connections to transfer context information from encoder to decoder \cite{Badrinarayanan2015Segnet,noh2015learning,bilin2018dense,ding2020se,olaf2015unet,yu2018learn,fu2019stack}. In \cite{Badrinarayanan2015Segnet,noh2015learning}, the spatial pooling indices of encoder are restored and then transferred in the upsampling process. U-Net \cite{olaf2015unet} designs a mirrored architecture for medical image segmentation, which copies the context features from encoder to decoder. RefineNet \cite{Guosheng2017RefineNet} utilizes a multi-path subnetwork to encode mulit-scale context, where the coarse semantic features are refined by fine-grained low-level features.  Yu \emph{et al.} \cite{yu2018learn} learn discriminatively contextual features and the additional edge clues in decoder stage. GCN \cite{chao2017large} investigates context cues using large convolution kernels to enlarge receptive fields. Ding \emph{et al.} \cite{ding2020se} propose a local contrasted context feature, which is selectively aggregated to distinguish foreground objects and background stuff. Beside capturing context information from encoder, some recent networks prefer to formulate context cues form decoder \cite{bilin2018dense,fu2019stack,tian2019dec}. For example, Fu \emph{et al.} \cite{fu2019stack} propose a cascaded EDN, where inter and intra connections are used to explore context information. Bilinski \emph{et al.} \cite{bilin2018dense} integrate full scale context to effectively perform information propagation through dense decoder skipped connections. Tian \emph{et al.}  \cite{tian2019dec} enable contextual aggregation by data-driven decoding. In contrast to these approaches that capture context in feature space, our method encodes contextual cues from the categorical perspective. Moreover, the BCIB employs bi-directional skipped connections, enabling our \emph{BiCANet} more powerful representation to explore context information.

\subsection{CANs}

CANs encode context information by integrating multi-scale features \cite{zhao2017pyramid,Chen2016deeplab,exploring2018lin,multi2016yu,conditional2015zheng,liu2018deep,wang2019deep}. A typical representative is dilated/atrous convolution \cite{Chen2016deeplab,yang2018dense,multi2016yu}, which investigates long-ranged context from surrounding positions with different dilated rate. For instance, DeepLab families \cite{Chen2016deeplab,Chen2017rethinking,chen2018enc} apply atrous convolution to produce larger size feature maps, which to some extend alleviates low-resolution features generated from traditional FCNs. Dilated-Net \cite{multi2016yu} appends several layers after the score map to embed the multi-scale context. Additionally, context prior can be captured from feature pyramid \cite{zhao2017pyramid,Chen2016deeplab} or image pyramid \cite{exploring2018lin,farabet2013learning}. For example, PSPNet \cite{zhao2017pyramid} designs a pyramid pooling block to aggregate multi-scale context. The atrous spatial pyramid pooling (ASPP) \cite{Chen2016deeplab} and its dense version \cite{yang2018dense} are proposed to capture the nearby spatial context using different dilation rates. HRNet \cite{wang2019deep} produces high resolution predictions by repeatedly aggregating contextual cues from multi-branched convolution streams. On the contrary, Farabet \emph{et al.} \cite{farabet2013learning} and Lin \emph{et al.} \cite{exploring2018lin} adopt multiple images with different resolution as input and fuse the corresponding features with different scales. An alternative approach embeds graph formulation into CNNs to aggregate long-ranged context information \cite{exploring2018lin,conditional2015zheng,liu2018deep,chandra2016fast}. For instance, CRF-RNN \cite{conditional2015zheng} and DSNet \cite{exploring2018lin} adopt mean field inference as recurrent layers for end-to-end training. Chen \emph{et al.} \cite{Chen2016deeplab} adopt CRF as post-process to produce more smooth segmentation results. Markov random field \cite{liu2018deep} is also utilized to capture long-ranged dependencies. In spite of achieving remarkable progress, the above approaches suffer from expensive computation, weakening the aggregation ability for limited scale features. Unlike these methods, our \emph{BiCANet} is implemented efficiently without any post-processing, leading to more powerful capacity to aggregate multi-scale context.

\subsection{AENs}

As local information often leads to classification ambiguity, AENs \cite{fu2019dual,zhang2019co,ding2019se,li2018pyr,zhao2018psanet,huang2019cc} prefer to learn global context information, which can be roughly divided into two categories: soft-attention \cite{hu2018se} and self-attention \cite{vas2017att}. The first category prefers to squeeze channel-wised global context by max pooling or average pooling. ParseNet \cite{liu2015parsenet} utilizes average pooling to encode global context for semantic segmentation. EncNet \cite{zhang2018con} calculates the pair-wise similarity between input and category-based codewords, which can be considered as channel-wised co-occurrent features. PANet \cite{li2018pyr} computes the channel-wise attention from high-level features to reweight low-level features in decoding pass. The second category, on the other hand, produces a powerful pixel-wise representation that calculates the correlation matrix between each pixel. For example, DANet \cite{fu2019dual} and OCNet \cite{yuan2018ocnet} enable a single feature for any pixel to interact with all other pixels. On the contrary, ACFNet \cite{zhang2019acf} only harvests the context information from the pixels that are associated with same semantic category. Ding \emph{et al.} \cite{ding2019se} embed object shape and layout context into self-attention. Non-local attention can be also considered as another kind of self-attention, which skillfully leverages the long-range dependencies for semantic segmentation \cite{wang2018non}. Zhang \emph{et al.} \cite{zhang2019co} design a co-occurrent feature model to formulate object co-occurrent context prior distribution. Zhu \emph{et al.} \cite{zhu2019asy} introduce an asymmetric pyramid non-local block to speed up computation. CCNet \cite{huang2019cc} decomposes a standard non-local module into two sequenced cross attention blocks. Although visual attention has gained substantial focuses by investigating global context, it is irrational to identify all individual pixels based on a single contextual scale. Conversely, \emph{BiCANet} predicts pixel semantic label from short-ranged to long-ranged, even to the global dependencies from the entire scene. Furthermore, MCFB enables us to selectively aggregate multi-scale features, which assigns appropriate contextual clues for different pixels.

\begin{figure*}[!t]
	\centerline{\includegraphics[width = 1.0\textwidth]{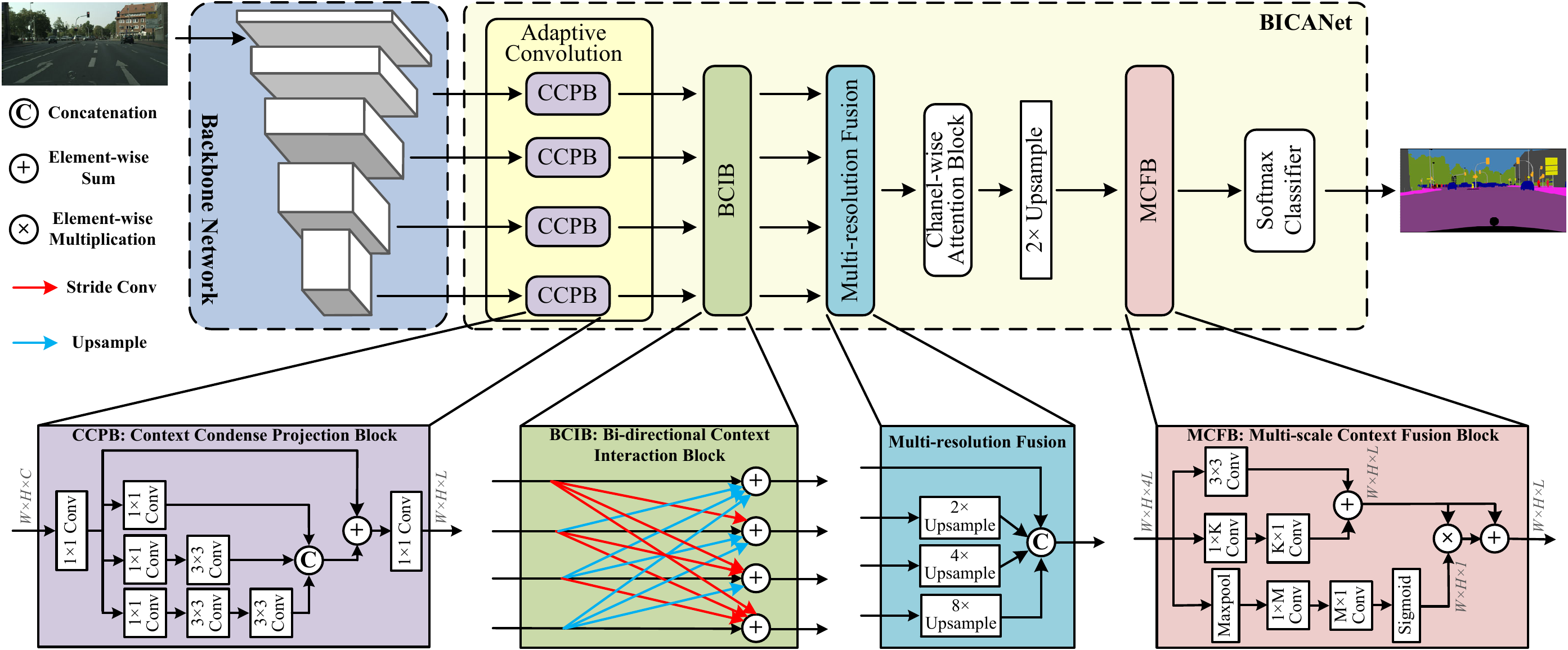}}
	\caption{The overall architecture of the proposed \emph{BiCANet}. The upper part shows the flowchart diagram of our method, from backbone to \emph{BiCANet}, and the lower part indicates the detailed structures of the main components of \emph{BiCANet}, including CCPB, BCIB, and MCFB, respectively. Please refer to text for more details. (Best viewed in color)} \label{fig:overall}
\end{figure*}

\section{Proposed Method}\label{sec:BiCANet}

We propose a new framework that captures context information from categorical perspective, where information from different resolutions and the potentially long-ranged dependencies are considered. Fig. \ref{fig:overall} shows the overall architecture of the proposed \emph{BiCANet}, which mainly consists of three parts: CCPB, BCIB, and MCFB. Immediately below, we elaborate on the details of these components.

\subsection{CCPB}

In spite of producing hierarchical feature representations, previous FCNs \cite{long2017fully,Chen2016deeplab,li2018pyr} inherently suffer from following limitations. Firstly, objects tend to be with different scales in images. However, traditional FCNs often have fixed receptive fields, resulting in the problem that objects substantially larger or smaller than receptive field may be fragmented or incorrectly classified \cite{Chen2016deeplab,Badrinarayanan2015Segnet}. Secondly, although low-level features reflect local statistics while high-level features stand for the overall property of an entire image, these abstracted features are always inaccurate and coarse, which is not beneficial for dense estimation problems. As a result, exploring context information based on these preliminary features may lead to poor segmentation results. In contrast, it is more robust to investigate context clues from the perspective of semantic categories, which is not apt to the influence from visual variety. In order to address these problems, the hierarchical features are condensed using a series of CCPBs, where each one adopts a split-transform-merge scheme to adaptively enlarge receptive fields, and simultaneously fulfills projection from feature space to semantic space. The detail structure of CCPB is illustrated in the purple block of Fig. \ref{fig:overall}. 

As can be seen, CCPB leverages the residual connections and multi-branch convolutions. The first one allows the convolutions to learn residual functions that facilitate training, while the second one adaptively enlarges receptive field to obtain more powerful representations. According to \cite{xie2017agg}, to ensure high computational efficiency, the feature channels before splitting and after merging are required to remain as equal as possible. Therefore, the input features $\mathbb{F} \in \mathbb{R}^{W \times H \times C}$ first undergo an $1 \times 1$ convolution to output features $\mathbb{F}' \in \mathbb{R}^{W \times H \times C'}$ whose channel numbers are small than $\mathbb{F}$, where $C$ and $C'$ ($C' < C$) denote channel numbers for $\mathbb{F}$ and $\mathbb{F'}$, respectively, and $W \times H$ indicates the feature resolution. Thereafter, $\mathbb{F}'$ is evenly split into $D$ branches for forthcoming transformations. Actually, $D$ can be considered as cardinality, controlling the number of transformations \cite{xie2017agg}. One may adopt arbitrary number of $D$ to make full use of different transformations, yet this is impractical as $D$ leverages the computational efficiency and model complexity of CCPB. In this paper, $D$ is set to 3, as it achieves best available trade-off between calculating efficiency and model complexity. Let $\mathcal{T}_i, i = \{1, 2, 3\}$ be an arbitrary transformation function, and $\bm{\theta}_i$ be the associated parameters. An $1 \times 1$ convolution is first used in each $\mathcal{T}_i$ to produce low dimensional embedding. In last two branches, these low dimensional features are fed into a series of $3 \times 3$ convolutions to achieve different scales of receptive fields. Then, the transformed features in all branches are combined using concatenation:
\begin{equation}\label{eq:Transforms}
\mathcal{F} (\mathbb{F}') = \bigodot_{i=1}^D \mathcal{T}_i(\mathbb{F}', \bm{\theta}_i)
\end{equation}
where $\bigodot$ denotes the concatenation operation. The aggregated transformation in Eqn. (\ref{eq:Transforms}) serves as the
residual function, which is helpful for the end-to-end training:
\begin{equation}
\tilde{\mathbb{F}} = \mathbb{F}' + \mathcal{F} (\mathbb{F}')
\end{equation}
where $\tilde{\mathbb{F}}$ is the condensed output. Finally, an $1 \times 1$ convolution is employed to project $\tilde{\mathbb{F}}$ into semantic space, yielding features $\hat{\mathbb{F}} \in \mathbb{R}^{W \times H \times L}$ whose channel number equals to the number of pre-defined categories.

The most similar residual architecture of CCPB is Inception families \cite{going2015szegedy,szegedy2016re}, also adopting split-transform-merge operation. The major differences, however, are two-fold: (1) We prune the maxpooling branch adopted in the preliminary version of Inception \cite{going2015szegedy}, which has been shown invalid in the gains of performance; (2) All Inception \cite{going2015szegedy} or Inception-ResNet modules \cite{szegedy2016re} prefer to utilize variable feature channels in different convolutions. On the contrary, our method shares the same feature channels among the multiple paths, requiring relative small extra efforts for designing each path.

In addition, CCPB appears to have a similar residual block to ResNext \cite{xie2017agg}, which involves branching and concatenating in the residual function. Our CCPB yet has the following major differences: (1) Unlike ResNext \cite{xie2017agg} that shares the same topology among multiple paths, the goal of CCPB lies in exploring different size of receptive fields, resulting in the diverse transformations designed for each path. (2) To facilitate computation, an $1 \times 1$ convolution is often used at the end of residual function \cite{xie2017agg}, producing equal feature channels for integration. In contrast, this is no further required as the same feature channels are always guaranteed due to the elegant architecture designed in CCPB.

\subsection{BCIB}

\begin{figure}[!t]
	\centerline{\includegraphics[width = 0.5\textwidth]{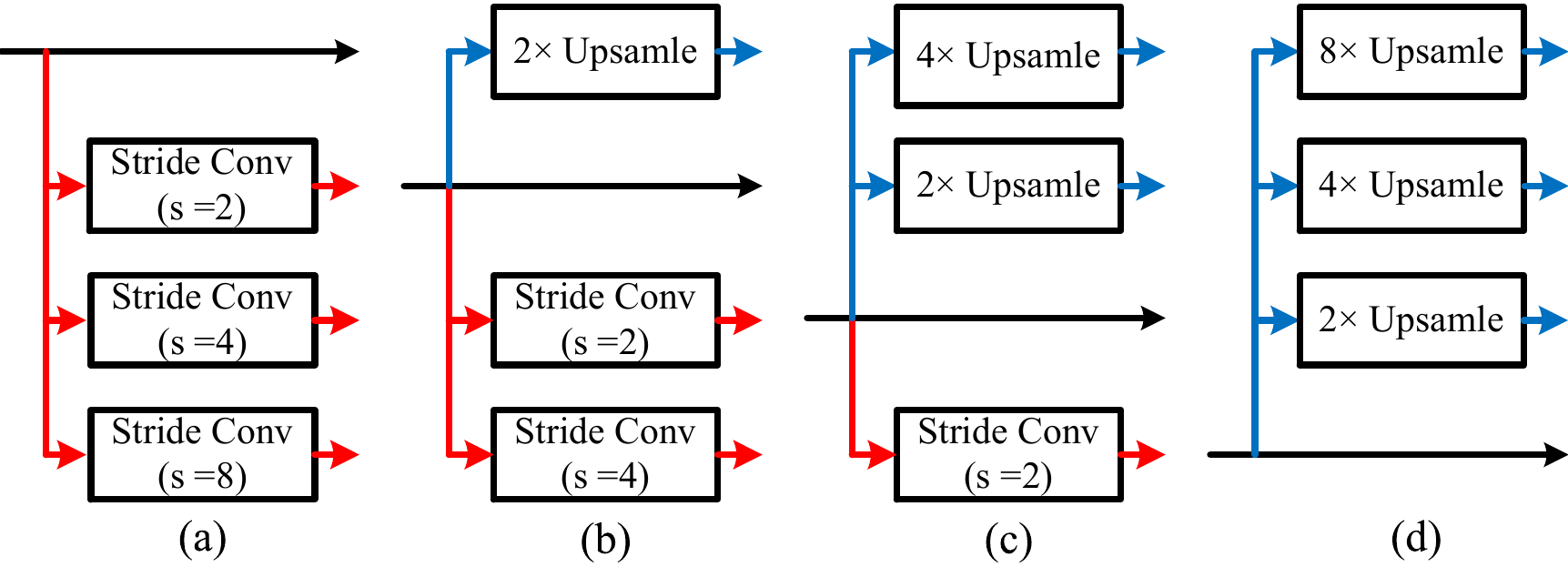}}
	\caption{Detailed structure of BCIB. As depicted in the green block of Fig. \ref{fig:overall}, the information flow of four branches is from (a) to (d) in BCIB, where black arrows denote identity mappings, red arrows indicate stride convolutions with different steps, and blue arrows are bilinear upsamplings with different ratios. (Best viewed in color).} \label{fig:BCIB}
\end{figure}

Due to the different resolutions in feature hierarchy, the feature maps are often resized and combined to encode multi-scale context information \cite{long2017fully,zhao2017pyramid,Badrinarayanan2015Segnet,noh2015learning}, which can be roughly divided into two categories: low-resolution representation and high-resolution recovering. The first category \cite{long2017fully} learns low-resolution representation to predict coarse segmentation maps, which are refined and augmented using contextual cues through feature pyramid \cite{zhao2017pyramid}. The second category gradually recovers the high-resolution representations from low level features using EDNs in a symmetric manner \cite{Badrinarayanan2015Segnet,noh2015learning} or asymmetric manner \cite{chen2018enc,lin2019refine}. Despite achieving promising performance, both of two categories share similar limitations: they do not make full use of context information with different scales. Rather than these representations, this section describes BCIB to enhance interactions of contextual cues, which adopts a multi-path parallel fashion through dense connections. 

The detail architecture of BCIB is illustrated in the green block of Fig. \ref{fig:overall}. As can be seen, there are four parallel paths of information flow, as we condense features from stage2 to stage5 of our backbone. At the end of each path, the context information from all paths are added, aggregating fine details from high-resolution representations and class-specific cues from low-resolution representations. Note except current path, the features from all other paths are required to be upsampled or downsampled, resulting in features with equal resolution for exact integration. More specifically, the upsampling or downsampling operations for each path are shown in Fig. \ref{fig:BCIB}. The simple bilinear upsamplings with different ratios are adopted to enlarge resolution, while stride convolutions with different steps are utilized to reduce resolution. Mathematically, from shallow to deep, let $\mathbb{F}_i$ be the input feature of $i^{th}$ path, where $i \in \{1, 2, 3, 4\}$, and $\mathcal{U}_m$ and $\mathcal{D}_n$ be the upsampling and downsampling operations for $m^{th}$ and $n^{th}$ paths, respectively. Then, the output features $\hat{\mathbb{F}}_i$ of  $i^{th}$ path, which combines the context information from all paths, are defined as:
\begin{equation}\label{eq:BCIB}
\hat{\mathbb{F}}_i = \mathbb{F}_i + \sum_{m = i + 1}^4 \mathcal{U}_m (\mathbb{F}_m) + \sum_{n = 1}^{i - 1} \mathcal{D}_n (\mathbb{F}_n, \bm{\theta}_n)
\end{equation}
where $\bm{\theta}_n$ is the associated parameters of stride convolution for $n^{th}$ path. From Eqn. (\ref{eq:BCIB}), beside making full use of context information from all paths that supplement each other to boost performance, another main merit of BCIB lies in the fact that the downsampling and upsampling operations can serve as a residual function between input and output features in each path, which is helpful to train BCIB in an end-to-end manner.

\subsection{MCFB}

The task of contextual formulation is to harvest surrounding information, which is always accomplished by enlarging the receptive field of convolution. In spite of producing high-level context features that represent entirety of an image, traditional DCNNs \cite{he2016res,going2015szegedy} are short for providing discriminative context information to assign categorical labels of local pixels, especially for those of inconspicuous and tiny objects. Lots of efforts have been devoted to capture coarse context (e.g., dilated convolution \cite{Chen2016deeplab,multi2016yu}) or dense context (e.g., GCN \cite{chao2017large} and RNN \cite{conditional2015zheng}) for semantic segmentation. These approaches, however, heavily depend on single scale context cues, lacking robust high-level contextual representations. Furthermore, due to the great visual variety of objects and stuff, indiscriminately harvesting context clues may result in harmful interference, especially under the case of cluttered surroundings. As a consequence, it is very hard to assign appropriate and discriminative context information for identifying local pixels.

\begin{figure}[!t]
	\centerline{\includegraphics[width = 0.5\textwidth]{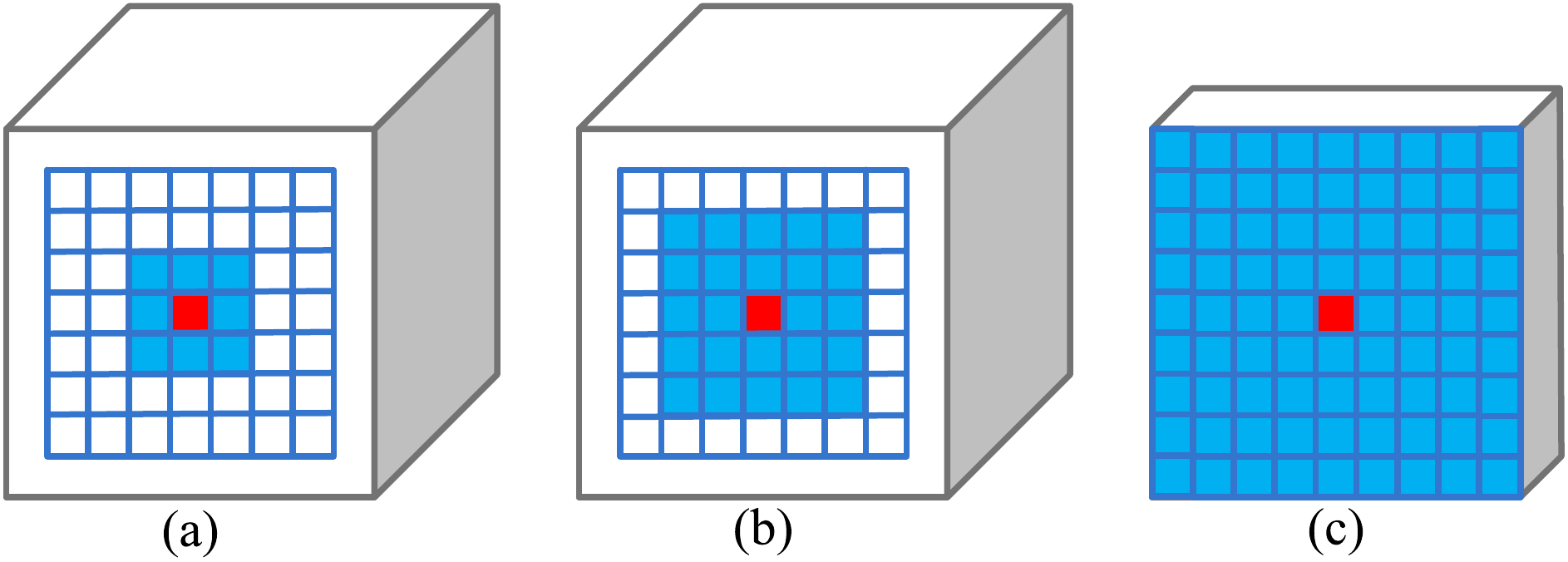}}
	\caption{Visualization of spatial context with different ranges. To identify semantic label of central red pixel, context information with different ranges (denoted as blue pixels) are used to reduce the classification ambiguities. In (a), the local context is captured using $3 \times 3$ convolution. In (b), long-ranged interactions are encoded using factorized convolutions (e.g., $K \times 1$ and $1 \times K$, where $K = 5$), which have the same receptive fields of $K \times K$ convolution, but with fewer parameters. Similar to (b), larger size factorized convolutions (e.g., $M \times 1$ and $1 \times M$, where $M = \max (W, H)$) are used to achieve similar receptive fields of $M \times M$ convolution, which are employed to explore global context of the entire scene, as illustrated in (c). Note, unlike long-ranged dependencies, the global context is explored in the squeezed feature maps. Please refer to text for more details. (Best viewed in color).} \label{fig:MCFB}
\end{figure}

Herein, it is essential to design tailored context features for different pixels. In addition, there exists richer semantic categories and their complex interactions \cite{shi2018hier}. Towards this end, this section proposes MCFB to address above problems, where high-level contextual features are appropriately assigned to different pixels. More specifically, as shown in Fig. \ref{fig:MCFB}, three scales of context cues are first investigated to capture local, long-ranged, and global interactions, respectively. After that, these contextual features are adaptively ensembled so that they supplement each other to make final discriminative decisions.

The detail architecture of MCFB is exhibited in the orange block of Fig. \ref{fig:overall}. As can be seen, the input features $\mathbb{F} \in \mathbb{R}^{W \times H \times 4L}$ pass through three branches, where each one corresponds to a case in Fig. \ref{fig:MCFB}. Mathematically, let $\mathcal{F}_s$, $\mathcal{F}_l$, and $\mathcal{F}_g$ be the contextual convolution of local dependencies, long-ranged interactions, and global context, respectively. Correspondingly, $\bm{\theta}_s$, $\bm{\theta}_l$, and $\bm{\theta}_g$ denote their respective parameters. As shown in Fig. \ref{fig:overall}, the first two branches output features $\mathbb{F}_{sl}$ by integrating local and long-ranged context:
\begin{equation}
\mathbb{F}_{sl} = \mathcal{F}_s (\mathbb{F}, \bm{\theta}_s) + \mathcal{F}_l(\mathbb{F}, \bm{\theta}_l)
\end{equation}
On the other hand, the final branch produces a spatial attention map $\mathbb{F}_g$ that grabs global context information. It first squeezes input features $\mathbb{F}$ into one channel using maxpooling, then a factorized convolution with large filter kernel size (e.g., $1 \times M$ and $M \times 1$) is utilized to harvest global dependencies from all pixels of the entire scene, which is normalized using a Sigmoid function $\mathcal{S}(\cdot)$:
\begin{equation}
\mathbb{F}_g = \mathcal{S} (\mathcal{F}_g (\emph{MaxPool}(\mathbb{F}), \bm{\theta}_g))
\end{equation}
Finally, the output features $\hat{\mathbb{F}}$ are selectively aggregated from $\mathbb{F}_{sl}$ and its counterpart reweighted by spatial attention map $\mathbb{F}_g$:
\begin{equation}
\hat{\mathbb{F}} = \mathbb{F}_{sl} + \mathbb{F}_g \otimes \mathbb{F}_{sl}
\end{equation}
where $\otimes$ stands for element-wise multiplication, and an identity mapping is used to leverage model training and multi-scale context fusion.

Surrounding context brings additional information, which is always useful for identifying pixels. The local dependencies, on the other hand, focus on partial information, yet often neglecting some essential clues provided by other parts. In contrast, aggregating multi-scale context via ensembling convolutions with different receptive fields forces MCFB to produce tailored features for different pixels, thus yielding more robust contextual representations.

\subsection{Network Architecture}

The overall architecture of our network is depicted in Fig. \ref{fig:overall}. In order to obtain high-quality semantic segmentation outputs and be consistent with the blocks of \emph{BiCANet} in connection pattern, we adopt ResNet-101, pre-trained on ImageNet \cite{deng2009imagenet}, as the backbone to abstract deep features. Following \cite{he2019ada,li2018pyr}, we remove the final fully connected layer and classification layer to ensure 2D representation that facilitates semantic segmentation. As a result, there are five stages in our backbone module, where each one has the resolution of $\frac{1}{2}$, $\frac{1}{4}$, $\frac{1}{8}$, $\frac{1}{16}$, and $\frac{1}{32}$ with respect to input image. Some previous works \cite{huang2019cc,zhang2019acf,zhu2019asy} prefer to employ holding-resolution version of ResNet-101 using dilation convolutions, where all the feature maps in the last three stages have the same spatial size. These methods, however, are at the cost of expensive computation and gridding artifacts that may degrade the performance. Note our proposed framework is flexible, as one can replace backbone with recent advances, such as ResNext \cite{xie2017agg} and DenseNet \cite{huang2017den}, to achieve better performance.

After we gather features from stage2 to stage5, an adaptive convolution is performed using a serious of CCPBs, which project abstracted features to categorical space. Motivated by \cite{zhao2017pyramid,fu2019stack}, we also add additional supervision to CCPBs, which is helpful for training the whole network. Furthermore, such supervision also suppresses the noises in the features from shallow layers, as it is beneficial to improve performance. As done for CCPBs, their outputs are fed into BCIB, allowing the information to flow and fuse within different levels. The fused features are thereupon concatenated with each other, avoiding the situation that BCIB could not produce accurate strengthened features. Note the features, which have different resolutions, are required to be upsampled to equal size for stacking. Thereafter, the stacked features undergo a channel-wise attention block using \cite{hu2018se}, and then are upsampled two times to match the resolution of the input image. Such features, full of rich contextual cues from different feature levels, serve as the input to MCFB, which then helps to explore the mutual correlations among pixels from local surroundings to global dependencies. Finally, the outputs of MCFB, which have predicted channel-wise semantic maps, later receive their supervisions from the ground truth maps.

\section{Experimental Results}\label{sec:Experimence}

In order to demonstrate the effectiveness of our method, we have conducted exhausted experiments on three widely-used semantic segmentation datasets: PASCAL VOC 2012 \cite{mark2015pascal}, Cityscapes \cite{Cordts2016the}, and ADE20K \cite{zhou2017sce}. In addition, we carry on a series of ablation studies to uncover the underlying impact of various components on the performance. Experimental results show that, compared with recent state-of-the-art approaches, our \emph{BiCANet} achieves superior performance on three datasets.

\subsection{Datasets and Evaluation Metrics}

\textbf{PASCAL VOC 2012 dataset} contains 21 object categories (20 foreground categories and one additional background class), which provides pixel-level annotation for each image. The original dataset has 1,464, 1,449, and 1,456 images for training, validation, and testing, respectively. Consistently with previous studies \cite{zhao2017pyramid,fu2019stack}, we augment the training set with extra annotated images provided by semantic boundaries \cite{Bharath2011semantic}, resulting in total 10,582 images for training.

\textbf{Cityscapes dataset} focuses on street scenes segmentation, and includes 30 object categories selected from 5 videos. This dataset has 5,000 high quality finely annotated images and 20,000 coarsely annotated images, where each image is shot on streets and of high-resolution ($2048 \times 1024$). Following \cite{Chen2016deeplab,fu2019dual,yang2018dense,lin2019refine}, only 19 classes are used for evaluation, and we only employ images with fine pixel-level annotations, resulting in 2,975 training, 500 validation and 1,525 testing images.

\textbf{ADE20K dataset} is a large-scale dataset used in ImageNet Scene Parsing Challenge 2016, containing up to 150 classes with a total of 1,038 image-level labels for diverse scenes. The categories include a large variety of objects and stuff. Similar to \cite{ding2020se,zhu2019asy,lin2019refine}, the dataset is divided into 20K/2K/3K images for training, validation, and testing, respectively. Unlike Cityscapes, both scenes and stuff are annotated in this dataset, resulting in more challenges for participated approaches.

\textbf{Evaluation Metric.} The performance is measured in terms of mean pixel intersection-over-union (mIoU) averaged across all semantic classes for all datasets. In addition, we also use pixel-wise accuracy and final score to evaluate segmenting results of ADE20K test dataset, where the first one is the ratio of correctly classified pixels with respect to ground truth, and the second is provided by the ADE20K organizers.

\subsection{Implementation Details}

\textbf{Training Objectives.} Following \cite{zhao2017pyramid,fu2019stack}, our training objective has two supervisions. The first one $\mathcal{L}_f$ is after the final output of our system, while the second one $\mathcal{L}_i$, $i = \{1, 2, 3, 4\}$, is at the output from each CCPB. Therefore, our loss function is composed by two cross entropy losses as
\begin{equation}\label{eq:loss}
\mathcal{L} = \mathcal{L}_f + \lambda \sum_i \mathcal{L}_i
\end{equation}
where $\lambda$ is a non-negative parameter that leverages the trade-off between two losses. For training $\mathcal{L}_f$, we employ online hard example mining \cite{shr2016tra}, which excels at coping with difficult cases. In our experiment, $\lambda$ is set to 0.1, empirically.

\textbf{Training Settings.} Our \emph{BiCANet} is implemented on the hardware platform of Intel Xeon E5-2680 server with NVIDIA 2080 Titan GPU. The software code is based on an open source repository for semantic segmentation \cite{abadi2016ten} using TensorFlow 2.0 framework \cite{all2020an}. The ResNet-101 backbone is pretrained on the ImageNet \cite{deng2009imagenet}, and then fine-tuned on three datasets to achieve best performance. Our \emph{BiCANet} is trained using the stochastic gradient descent algorithm \cite{Bottou2010sgd}, where the initial learning rate is set to $2 \times 10^{-2}$ for ADE20K and $10^{-2}$ for rest datasets, together with momentum and weight decay, which are set to 0.99 and $10^{-4}$, respectively. Following \cite{Chen2016deeplab}, we employ a ``poly'' learning rate policy in training process, where the initial learning rate is multiplied by $(1 - \frac{iter}{max\_iter})^{power}$ with $power = 0.9$. To augment training data, we first randomly crop out high-resolution patches with resolution of $769 \times 769$ from original images as the inputs for Cityscapes \cite{zhao2017pyramid,Chen2017rethinking}. While for the other two datasets, as well as \cite{zhao2017pyramid,zhang2018con}, the cropped resolutions are set to $480 \times 480$ and $520 \times 520$, respectively. Moreover, we also apply random scaling in the range of [0.5, 2.0], aspect ratios in the range of [0.7, 1.5], horizontal flip, and left-right flip as additional data augmentation methods for all datasets. Batch size is set to 8 for all datasets, where we favor a large minibatch size to make full use of the GPU memory. As discussed before, we also employ the auxiliary loss $\mathcal{L}_i$ and online hard example mining strategy \cite{shr2016tra} in all experiments, as their benefits for improving performance have been clearly discussed in \cite{zhao2017pyramid}. Finally, our model is trained using 200 epochs for all datasets. In inference, the testing results on three datasets were submitted to the official online servers for evaluation.

\textbf{Baselines.} In order to show the advantages of \emph{BiCANet}, we selected 18 state-of-the-art networks as baselines for comparison, including FCN-8s \cite{long2017fully}, DeepLabV3 \cite{chen2018enc}, CRF-RNN \cite{conditional2015zheng}, DliateNet \cite{multi2016yu}, OCNet \cite{yuan2018ocnet}, SegNet \cite{Badrinarayanan2015Segnet}, DenseASPP \cite{yang2018dense}, GCN \cite{chao2017large}, PSPNet \cite{zhao2017pyramid}, DANet \cite{fu2019dual}, EncNet \cite{zhang2018con}, HRNetV2 \cite{wang2019deep}, CCNet \cite{huang2019cc}, SDNet \cite{fu2019stack}, MPDNet \cite{ding2020se}, RefineNet \cite{lin2019refine}, CFNet \cite{zhang2019co}, and APCNet \cite{he2019ada}.

\begin{figure*}[!t]
	\centerline{\includegraphics[width = 1.0\textwidth]{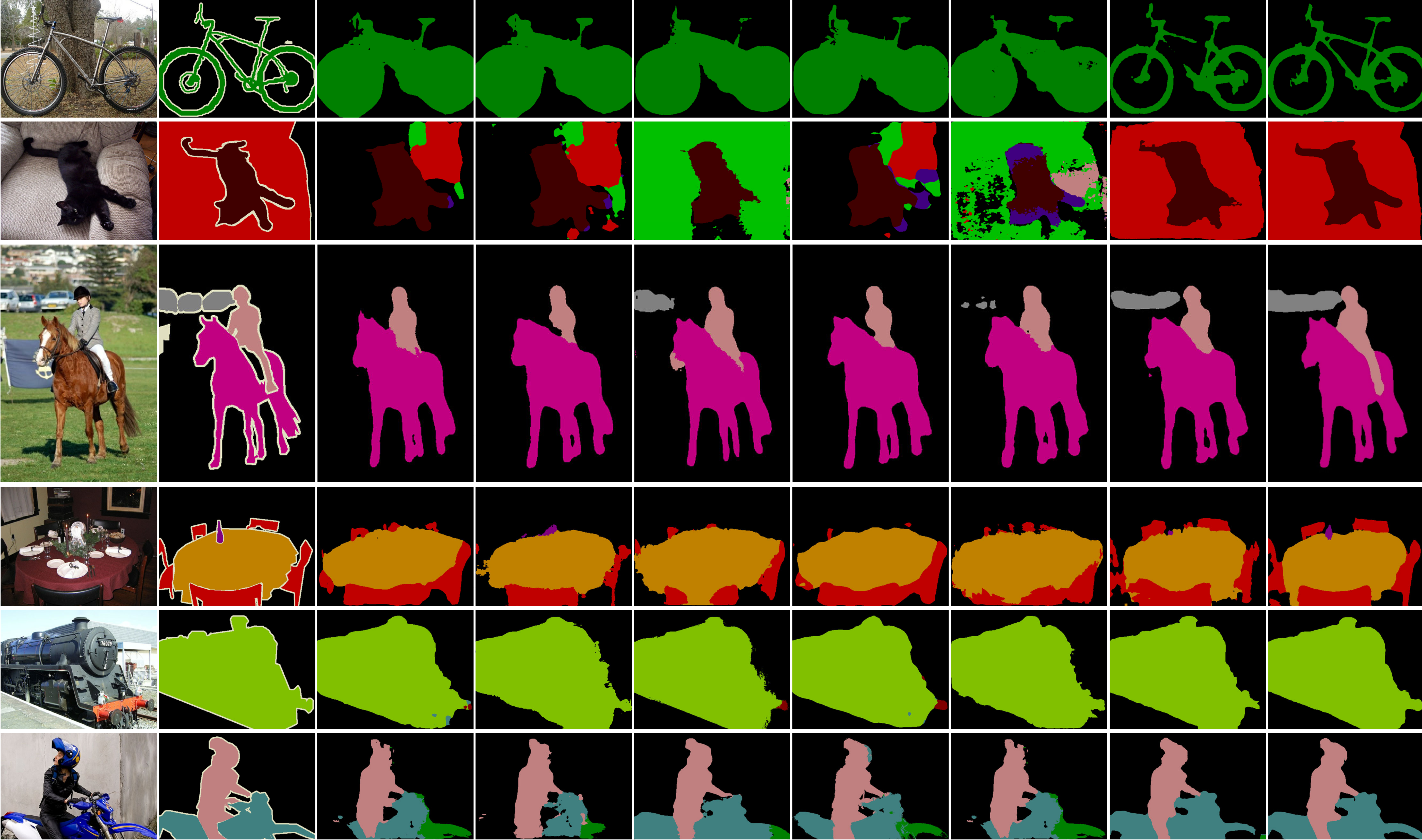}}
	\caption{Some visual comparisons on PASCAL VOC 2012 \textit{validation} dataset. From left to right are input images, the corresponding ground truth, segmentation outputs from SegNet \cite{Badrinarayanan2015Segnet}, FCN-8s \cite{long2017fully}, CRF-RNN \cite{conditional2015zheng}, RefineNet \cite{lin2019refine}, PSPNet \cite{zhao2017pyramid}, DeepLabV3 \cite{chen2018enc}, and our \emph{BiCANet}. It is evident that, compared with state-of-the-art baselines, \emph{BiCANet} produces more accurate predictions with better delineated object boundaries and shapes. (Best viewed in color)} \label{fig:VOCResults}
\end{figure*}

\begin{table*}[!t]
	\tabcolsep 1.0mm \caption{Individual category results on the PASCAL VOC 2012 \textit{test} set in terms of mIOU scores. The bold number indicates the best performance among all approaches for each category.}
	\begin{center}
		\begin{tabular}{c|cccccccccccccccccccc|c}
			\toprule
			Method  &{aero}  &{bike}  &{bird}  &{boat}  &{bottle}  &{bus}  &{car}  &{cat}  &{chair}  &{cow}  &{table}  &{dog}  &{horse}  &{mbk}  &{person}  &{plant}  &{sheep}  &{sofa}  &{train}  &{tv}  &{mIoU}\\
			\midrule \midrule
			SegNet \cite{Badrinarayanan2015Segnet}     &74.5 &30.6 &61.4 &50.8 &49.8 &76.2 &64.3 &69.7 &23.8 &60.8 &54.7 &62.0 &66.4 &70.2 &74.1 &37.5 &63.7 &40.6 &67.8 &53.0 &59.1\\
			FCN-8s \cite{long2017fully}        &76.8 &34.2 &68.9 &49.4 &60.3 &75.3 &74.7 &77.6 &21.4 &62.5 &46.8 &71.8 &63.9 &76.5 &73.9 &45.2 &72.4 &37.4 &70.9 &55.1 &62.2\\
			CRF-RNN \cite{conditional2015zheng}    &87.5 &39.0 &79.7 &64.2 &68.3 &87.6 &80.8 &84.4 &30.4 &78.2 &60.4 &80.5 &77.8 &83.1 &80.6 &59.5 &82.8 &47.8 &78.3 &67.1 &72.0\\
			HRNetV2 \cite{wang2019deep}    &93.8 &43.5 &84.8 &63.9 &82.4 &92.8 &91.0 &93.8 &45.6 &88.0 &61.4 &90.0 &90.2 &88.0 &88.1 &66.8 &91.1 &53.3 &87.1 &74.4 &79.3\\
			RefineNet \cite{lin2019refine}  &95.0 &73.2 &93.5 &78.1 &84.8 &95.6 &89.8 &94.4 &43.7 &92.0 &77.2 &90.8 &93.4 &88.6 &88.1 &70.1 &92.9 &64.3 &87.7 &78.8 &83.8\\
			PSPNet \cite{zhao2017pyramid}     &95.8 &72.7 &\textbf{95.0} &78.9 &84.4 &94.7 &92.0 &95.7 &43.1 &91.0 &\textbf{80.3} &91.3 &96.3 &92.3 &90.1	&71.5 &94.4	&66.9 &88.8	&\textbf{82.0} &85.4\\
			DeepLabV3 \cite{chen2018enc}  &96.4 &76.6 &92.7 &77.8 &\textbf{87.6} &96.7 &90.2 &95.4 &47.5 &93.4 &76.3 &91.4 &\textbf{97.2} &91.0 &92.1	&71.3 &90.9 &\textbf{68.9} &90.8 &79.3 &85.7\\
			EncNet \cite{zhang2018con}     &95.3 &76.9 &94.2 &\textbf{80.2} &85.2 &96.5 &90.8 &96.3 &\textbf{47.9} &\textbf{93.9} &80.0 &92.4 &96.6 &90.5 &91.5	&70.8 &93.6 &66.5 &87.7 &80.8 &85.9\\	
			\midrule
			BiCANet    &\textbf{97.1} &\textbf{79.0} &92.8 &76.2 &82.9 &\textbf{97.8} &\textbf{93.6} &\textbf{98.4} &45.1 &93.5 &75.3 &\textbf{93.9} &96.2 &\textbf{92.6} &\textbf{92.4}	&\textbf{79.3} &\textbf{95.8} &\textbf{68.9} &\textbf{91.1} &80.3 &\textbf{86.7}\\
			\bottomrule
		\end{tabular}
	\end{center}\label{tab:VOC}
\end{table*}

\begin{table}[!t]
	\tabcolsep 3.0mm \caption{Comparison with state-of-the-art methods on the PASCAL VOC 2012 \textit{test} set in terms of mIOU scores. The bold number indicates the best performance among all approaches.}
	\begin{center}
		\begin{tabular}{c|ccc}
			\toprule
			Method &Year &Backbone  &mIoU\\
			\midrule
			SegNet  \cite{Badrinarayanan2015Segnet}  &TPAMI2017  	&VGG-16    &59.1\\
			FCN-8s  \cite{long2017fully}     &TPAMI2017	&VGG-16  &62.2\\
			CRF-RNN \cite{conditional2015zheng}  &ICCV2015  &VGG-16      &72.0\\
			HRNetV2 \cite{wang2019deep}  &arXiv2019  &-	&82.6\\
			DANet \cite{fu2019dual}      &CVPR2019   &Dilated-ResNet-101	&82.6\\
			SDNet \cite{fu2019stack} &TIP2019	&DenseNet-161	&83.5 \\
			GCN \cite{chao2017large}       &CVPR2017		&ResNet-152	&83.6\\
			RefineNet \cite{lin2019refine}  &TPMAI2019   &ResNet-152	&83.8\\
			CFNet \cite{zhang2019co} &CVPR2019	&Dilated-ResNet-101		&84.2\\  
			APCNet \cite{he2019ada}		&CVPR2019		&Dilated-ResNet-101		&84.2   \\
			PSPNet \cite{zhao2017pyramid}   &CVPR2016  &Dilated-ResNet-101	&85.4\\
			DeeplabV3 \cite{chen2018enc}   &ECCV2018	&ResNet-101	&85.7\\
			EncNet \cite{zhang2018con}     &CVPR2018   &Dilated-ResNet-101	&85.9\\
			\midrule 
			BiCANet    &-	&ResNet-101		&\textbf{86.7}\\
			\bottomrule
		\end{tabular}
	\end{center}\label{tab:VOCALL}
\end{table}

\subsection{Comparisons with State-of-the-art}

This section reports the results of our method, and compares with sate-of-the-art networks.

\subsubsection{Results on PASCAL VOC 2012}

Tables \ref{tab:VOC} and \ref{tab:VOCALL} report the quantitative results of each individual category using our \emph{BiCANet}, and compare it with previous approaches on Pascal VOC 2012 \emph{test} set. Our \emph{BiCANet} outperforms previous state-of-the-art networks, achieving best performance with 86.7\% mIoU accuracy. Although our method does not involve any post-processing, it is intriguing that \emph{BiCANet} is superior to the existing methods \cite{chen2018enc} that employ CRF as post-processing. This indicates our \emph{BiCANet} is able to adaptively capture appropriate context information to improve performance. From Table \ref{tab:VOC}, it is observed that our method obtains best mIoU scores on 12 out of the 20 object categories, achieving remarkable improvement than the second-ranked method on some categories (e.g., 7.8\% for ‘Plant’ and 2.1\% for ‘Motobike’). From Table \ref{tab:VOCALL}, it is discovered that methods employing VGG-based backbones are ranked the lowest. The approaches that utilize Dilated-ResNet-101 as backbones, such as PSPNet \cite{zhao2017pyramid}, APCNet \cite{he2019ada}, and DANet \cite{fu2019dual}, always achieve better results. It is also interesting that, in spite of using ResNet-101 as backbones, our \emph{BiCANet} outperforms GCN \cite{chao2017large} and SDNet \cite{fu2019stack}, which employ backbones with more deeper architecture, e.g., ResNet-152 and DenseNet-161. Moreover, the performance could be further boosted by pre-training our backbone on MS COCO dataset, as is shown in \cite{zhang2019co,zhang2018con}.

Fig. \ref{fig:VOCResults} shows the qualitative results of some visual examples on PASCAL VOC 2012 \emph{validation} dataset. Each example shows both the original image and the color coded segmentation output. It is evident that, compared with state-of-the-art baselines, our \emph{BiCANet} produces more smooth outputs and more accurate predictions with delineated object boundaries and shapes, which is consistent with the results reported in Tables \ref{tab:VOC} and \ref{tab:VOCALL}. For instance, the tires of bicycle in the first example, the background sofa in the second example, and the background cars and the foreground human legs in the third example. Moreover, our method is very effective for correctly classifying tiny objects, such as the small bottle on the table, which is omitted by other baselines.

\begin{figure*}[!t]
	\centerline{\includegraphics[width = 1.0\textwidth]{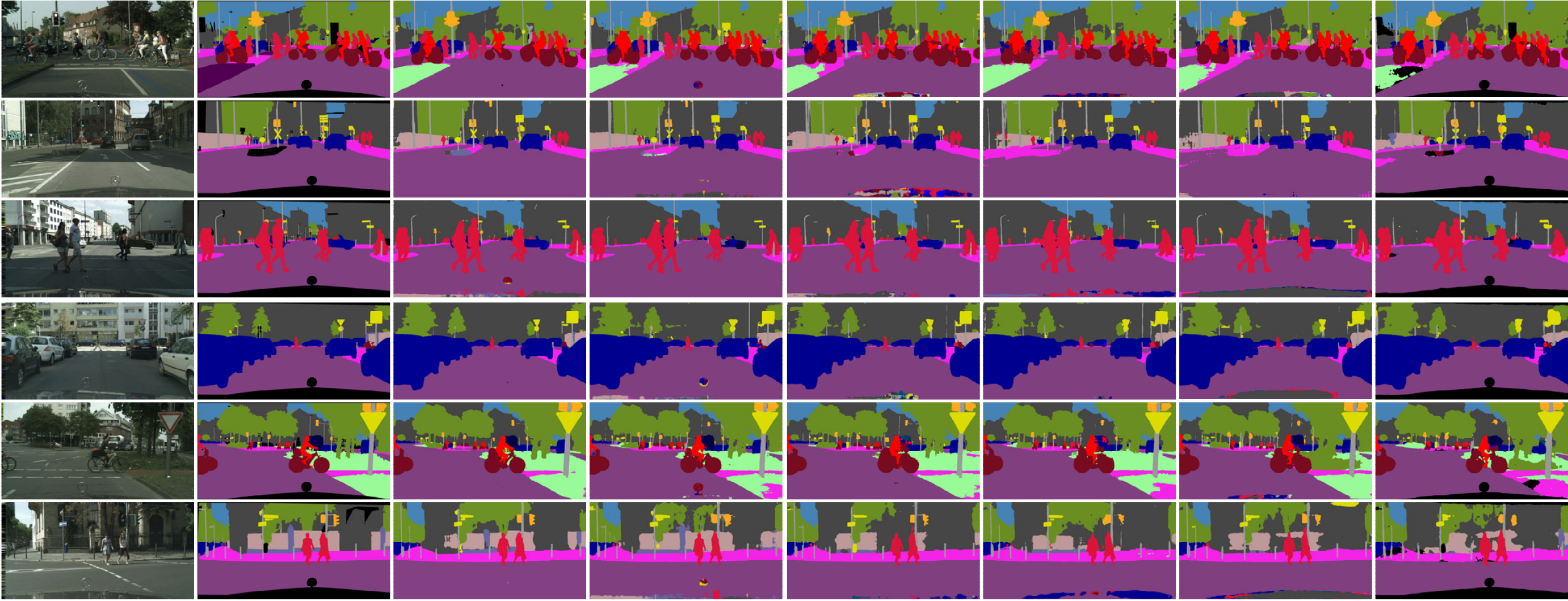}}
	\caption{Some visual comparisons on Cityscapes \textit{validation} dataset. From left to right are input images, the corresponding ground truth, segmentation outputs from our \emph{BiCANet}, DeepLabV3 \cite{chen2018enc}, PSPNet \cite{zhao2017pyramid}, RefineNet \cite{lin2019refine}, CRF-RNN \cite{conditional2015zheng}, and FCN-8s \cite{long2017fully}. It is evident that, compared with state-of-the-art baselines, our \emph{BiCANet} produces more accurate predictions with better delineated object boundaries and shapes. (Best viewed in color).} \label{fig:City}
\end{figure*}

\begin{table*}[!t]
	\tabcolsep 1.4mm \caption{Individual category results on the Cityscapes \textit{test} set in terms of mIOU scores. The bold number indicates the best performance among all approaches for each category.}
	\begin{center}
		\begin{tabular}{c|ccccccccccccccccccc|c}
			\toprule
			Method &{Roa}  &{Sid}  &{Bui}  &{Wal}  &{Fen}  &{Pol}  &{TLi}  &{TSi}  &{Veg}  &{Ter}  &{Sky}  &{Ped}  &{Rid}  &{Car}  &{Tru}  &{Bus}  &{Tra}  &{Mot}  &{Bic}  &{mIoU} \\
			\midrule \midrule
			SegNet \cite{Badrinarayanan2015Segnet}    &96.4 &73.2 &84.0 &28.5 &29.0 &35.7 &39.8 &45.2 &87.0 &63.8 &91.8 &62.8 &42.8 &89.3 &38.1 &43.1 &44.2 &35.8 &51.9 &57.0  \\ 				
			CRF-RNN \cite{conditional2015zheng}   &96.3 &73.9 &88.2 &47.6 &41.3 &35.2 &49.5 &59.7 &90.6 &66.1 &93.5 &70.4 &34.7 &90.1 &39.2 &57.5 &55.4 &43.9 &54.6 &62.5  \\		
			FCN-8s \cite{long2017fully} 	  &97.4 &78.4 &89.2 &34.9 &44.2 &47.4 &60.1 &65.0 &91.4 &69.3 &93.9 &77.1 &51.4 &92.6 &35.3 &48.6 &46.5 &51.6 &66.8 &65.3  \\				
			RefineNet \cite{lin2019refine} &98.2 &83.3 &91.3 &47.8 &50.4 &56.1 &66.9 &71.3 &92.3 &70.3 &94.8 &80.9 &63.3 &94.5 &64.6 &76.1 &64.3 &62.2 &70.0 &73.6  \\
			PSPNet \cite{zhao2017pyramid}    &98.7 &86.9 &93.5 &58.4 &\textbf{63.7} &67.7 &76.1 &80.5 &93.6 &72.2 &95.3 &86.8 &71.9 &96.2 &77.7 &91.5 &83.6 &70.8 &77.5 &81.2  \\
			DeepLabV3 \cite{chen2018enc} &98.6 &86.2 &93.5 &55.2 &63.2 &70.0 &77.1 &81.3 &93.8 &72.3 &95.9 &87.6 &73.4 &96.3 &75.1 &90.4 &85.1 &72.1 &78.3 &81.3  \\
			DANet \cite{fu2019dual}     &98.6 &86.1 &93.5 &56.2 &63.3 &69.7 &77.3 &81.3 &93.9 &72.9 &95.7 &87.3 &72.9 &96.2 &76.8 &89.5 &86.5 &72.2 &78.2 &81.5  \\
			HRNetV2 \cite{wang2019deep}   &\textbf{98.8} &87.9 &93.9 &61.3 &63.1 &\textbf{72.1} &\textbf{79.3} &\textbf{82.4} &94.0 &73.4 &96.0 &\textbf{88.5} &75.1 &\textbf{96.5} &72.5 &88.1 &79.9 &73.1 &79.2 &81.8  \\			
			\midrule
			BiCANet &98.7 &\textbf{89.0} &\textbf{94.5} &\textbf{62.6} &63.6 &66.4 &75.6 &79.6 &\textbf{94.6} &\textbf{73.5} &\textbf{96.4} &87.2 	&\textbf{75.3} &94.1 &\textbf{79.3} &\textbf{91.8} &\textbf{86.7} &\textbf{73.2} &\textbf{79.6} &\textbf{82.4} \\
			\bottomrule
		\end{tabular}
	\end{center}\label{tab:City}
\end{table*}

\subsubsection{Results on Cityscapes}

\begin{table}[!t]
	\tabcolsep 3.0mm \caption{Comparison with state-of-the-art methods on the Cityscapes \textit{test} set in terms of mIOU scores. The bold number indicates the best performance among all approaches.}
	\begin{center}
		\begin{tabular}{c|ccc}
			\toprule
			Method &Year &Backbone  &mIoU\\
			\midrule
			SegNet  \cite{Badrinarayanan2015Segnet}  &TPAMI2017  	&VGG-16    &57.0\\			
			CRF-RNN \cite{conditional2015zheng}  &ICCV2015  &VGG-16      &62.5\\
			FCN-8s  \cite{long2017fully}     &TPAMI2017	&VGG-16  &65.3\\			
			RefineNet \cite{lin2019refine}  &TPMAI2019   &ResNet-101	&73.6\\
			GCN \cite{chao2017large}       &CVPR2017		&ResNet-152	&76.9\\
			CFNet \cite{zhang2019co} &CVPR2019	&Dilated-ResNet-101		&79.6\\  
			DenseASPP \cite{yang2018dense}		&CVPR2018		&ResNet-101		&80.6\\
			PSPNet \cite{zhao2017pyramid}   &CVPR2016  &Dilated-ResNet-101	&81.2\\
			DeeplabV3 \cite{chen2018enc}   &ECCV2018	&ResNet-101	&81.3\\
			CCNet \cite{huang2019cc} &ICCV2019	&Dilated-ResNet-101		&81.4\\  
			DANet \cite{fu2019dual}      &CVPR2019   &Dilated-ResNet-101	&81.5\\
			HRNetV2 \cite{wang2019deep}  &arXiv2019  &-	&81.8\\
			\midrule 
			BiCANet    &-	&ResNet-101		&\textbf{82.4}\\
			\bottomrule
		\end{tabular}
	\end{center}\label{tab:CityAll}
\end{table}

In this section, we demonstrate our method scales nicely on street scene dataset Cityscapes. Table \ref{tab:City} and \ref{tab:CityAll} report the quantitative results for each individual category of \emph{BiCANet}, and compare it with previous state-of-the-art baselines. Our method also has superior performance than state-of-the-art baselines, achieving 82.4\% mIoU among all categories. Compared with second-rank approach HRNetV2 \cite{wang2019deep} using more complicated network architecture to maintain high-resolution output, our \emph{BiCANet} is easy to implement, improving 0.6\% mIoU improvement. From Table \ref{tab:City}, it is observed that our method obtains best mIoU scores on 12 out of the 19 object categories. Particularly, some classes have a major boost in performance, including 1.1\% for ‘Sidewalk’, 1.3\% for ‘Wall’, and 1.6\% for ‘Truck’.

Fig. \ref{fig:City} shows some visual examples of obtained semantic segmentation output. It is evident that our network is not only robust to the categories with great appearance variety, i.e., ‘Building’ and ‘Car’, but also sketches well multi-scale objects, i.e., ‘Signs’ and ‘Pedestrians’. It is also discovered that \emph{BiCANet} produces more accurate predictions for different objects and regions, which are always misclassified by other baselines, such as cross ‘Sign’ in the second example, ‘Sidewalk’ in the fifth example, and ‘Tree’ in the final example. All the results on this dataset show that our method can capture more accurate context information and learn better adaptive-ranged spatial dependencies.

\subsubsection{Results on ADE20K}

\begin{figure*}[!t]
	\centerline{\includegraphics[width = 1.0\textwidth]{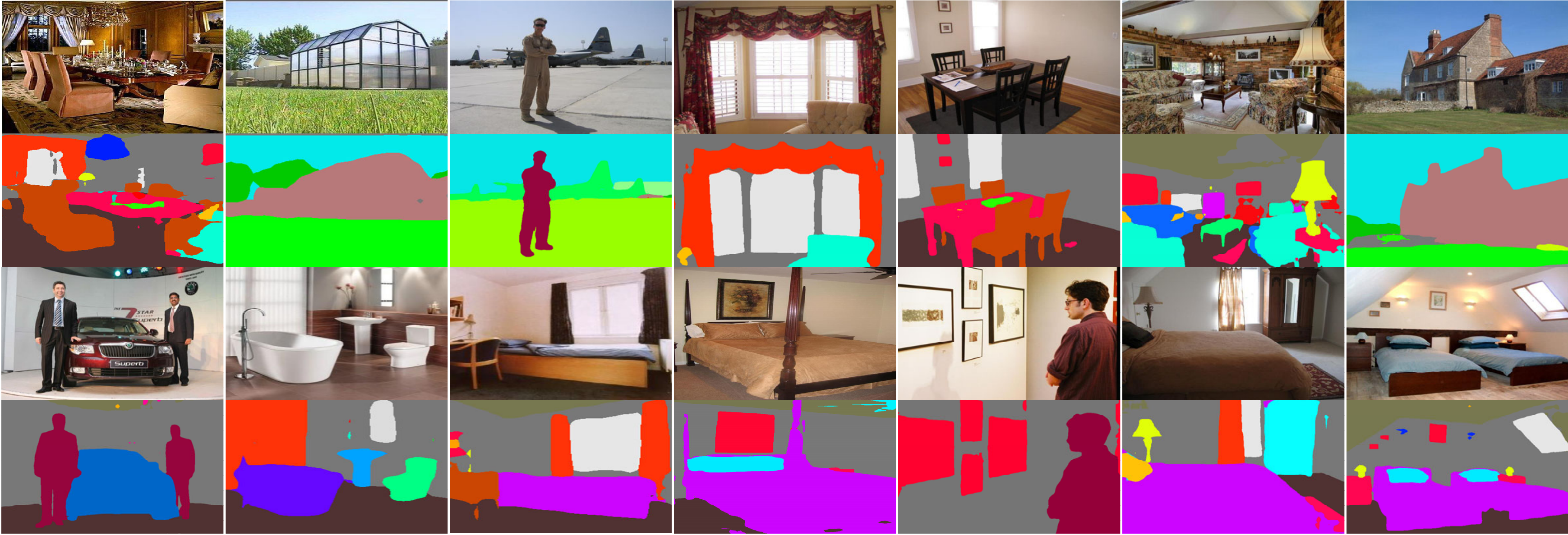}}
	\caption{Some visual examples of \emph{BiCANet} outputs on ADE20K \textit{validation} dataset. (Best viewed in color)} \label{fig:ADE}
\end{figure*}

\begin{table}[!t]
	\tabcolsep 2.5mm \caption{Comparison with state-of-the-art methods on the ADE20K \textit{test} set in terms of mIOU scores. The bold number indicates the best performance among all approaches.}
	\begin{center}
		\begin{tabular}{c|cccc}
			\toprule
			Method  &Year &PixAcc &mIoU &Final Score\\
			\midrule
			SegNet \cite{Badrinarayanan2015Segnet} &TPAMI2017 &64.03 &17.54 &0.4097   \\	
			FCN-8s \cite{long2017fully}  &TPAMI2017  &64.77 &24.83	&0.4480      \\
			DliateNet \cite{multi2016yu} &ICLR2016 &65.41	&25.99	&0.4558  \\			
			OCNet \cite{yuan2018ocnet} &arXiv2018   &73.61 &38.07 &0.5584  \\
			PSPNet \cite{zhao2017pyramid} &CVPR2016 &72.92	&38.13	&0.5538     \\
			EncNet \cite{zhang2018con} &CVPR2018 &73.14	&38.17	&0.5567    \\
			\midrule
			BiCANet &-	&\textbf{73.90}	&\textbf{38.66}	&\textbf{0.5588}   \\
			\bottomrule
		\end{tabular}
	\end{center}\label{tab:ADE}
\end{table}

In Table \ref{tab:ADE}, \emph{BiCANet} achieves 73.90\%, 38.66\%, and 0.5588 of pixel-wise accuracy, mIoU, and final score, respectively, which surpass PSPNet \cite{zhao2017pyramid} (1st place in 2016) and all other baselines. Particularly, the proposed method significantly outperforms the baseline SegNet \cite{Badrinarayanan2015Segnet}, improving 9.87\%, 21.12\%, and 0.1470 in terms of three evaluation metrics. In Fig. \ref{fig:ADE}, we also show some visual examples of original images and their corresponding color coded segmentation outputs. Consistent with the results on PASCAL VOC 2012 and Cityscapes, \emph{BiCANet} shows its outstanding ability on identifying small objects, such as ‘Light’, ‘Book’, and ‘Sign’.

\subsection{Ablation Study}

To understand the underlying behavior of our \emph{BiCANet}, this section reports the results of a series of ablation studies. Note all the experiments are evaluated on \emph{validation} set.

\subsubsection{Ablation study for components of BiCANet}

Table \ref{tab:ComponentPASCAL} and \ref{tab:ComponentCity} present ablation studies on PASCAL VOC 2012 and Cityscapes datasets, which quantify the influence of three components: CCPB, BCIB, and MCFB, as discussed earlier. This experiment shows that each of these components consistently improve the performance. It is also observed that, among all components, MCFB significantly improves the performance, e.g., 3.31\% and 2.63\% using ResNet-50 and ResNet-101 backbones, respectively, on PASCAL VOC 2012 dataset. We also observe similar results in Table \ref{tab:ComponentCity} on Cityscapes dataset. Generally, deeper initialization backbones, e.g., ResNet-101, always lead to better performance.

\begin{table}[!t]
	\tabcolsep 4.0mm \caption{Ablation experiments for analyzing the contributions of different components on PASCAL VOC 2012 \emph{validation} set. The bold number indicates the best performance in terms of mIoU.}
	\begin{center}
		\begin{tabular}{c|ccc|c}
			\toprule
			Backbone  &CCPB &BCIB &MCFB &mIoU\\
			\midrule
			ResNet-50 &~		 	&~			&~		 	&75.50   \\	
			ResNet-50 &\checkmark  	&~			&~       	&77.65  \\
			ResNet-50 &\checkmark	&\checkmark	&~	  	 	&78.41  \\			
			ResNet-50 &\checkmark   &\checkmark &\checkmark &\textbf{81.72}  \\
			\midrule
			ResNet-101 &~		 	&~			&~		 	&78.32   \\	
			ResNet-101 &\checkmark 	&~			&~       	&79.15  \\
			ResNet-101 &\checkmark	&\checkmark	&~	  	 	&80.70  \\			
			ResNet-101 &\checkmark  &\checkmark &\checkmark &\textbf{83.33}  \\
			\bottomrule
		\end{tabular}
	\end{center}\label{tab:ComponentPASCAL}
\end{table}

\begin{table}[!t]
	\tabcolsep 4.0mm \caption{Ablation experiments for analyzing the contributions of different components on Cityscapes \emph{validation} set. The bold number indicates the best performance in terms of mIoU.}
	\begin{center}
		\begin{tabular}{c|ccc|c}
			\toprule
			Backbone  &CCPB &BCIB &MCFB &mIoU\\
			\midrule
			ResNet-50 &~		 	&~			&~		 	&70.03   \\	
			ResNet-50 &\checkmark  	&~			&~       	&72.46  \\
			ResNet-50 &\checkmark	&\checkmark	&~	  	 	&75.93  \\			
			ResNet-50 &\checkmark   &\checkmark &\checkmark &\textbf{78.11}  \\
			\midrule
			ResNet-101 &~		 	&~			&~		 	&72.54   \\	
			ResNet-101 &\checkmark 	&~			&~       	&73.68  \\
			ResNet-101 &\checkmark	&\checkmark	&~	  	 	&75.45  \\			
			ResNet-101 &\checkmark  &\checkmark &\checkmark &\textbf{79.90}  \\
			\bottomrule
		\end{tabular}
	\end{center}\label{tab:ComponentCity}
\end{table}

\subsubsection{Ablation study for different backbones}

Different pre-trained backbones have been shown their diversity power to represent large scale visual data in previous literature. To further analyze \emph{BiCANet}, we conduct experiments on PASCAL VOC 2012 and Cityscapes datasets using different backbones. In particular, with the same setting, we test \emph{BiCANet} with different recent backbones, such as VGG-Net \cite{long2017fully}, ResNet \cite{he2016res}, ResNext \cite{xie2017agg} and DenseNet \cite{huang2017den}. As reported in Table \ref{tab:backbone}, using backbones that have more powerful representation ability often leads to an improvement in performance. Even so, we still employ ResNet-101 as backbone for fair comparison with most state-of-the-art networks. Note compared with VGG-16 backbones, using ResNet-101 achieves remarkable mIoU improvement of 5.78\% and 5.58\% on two datasets, respectively.

\begin{table}[!t]
	\tabcolsep 0.6mm \caption{Ablation experiments using different pre-trained backbones in our \emph{BiCANet}. The bold number indicates the best performance in terms of mIoU.}
	\begin{center}
		\begin{tabular}{c|cccc}
			\toprule
			Dataset &VGG-16 &ResNet-101  &ResNext-101	&DenseNet-161\\
			\midrule
			PASCAL VOC 2012   &77.55  	&83.33    &85.15  	&\textbf{85.47}\\			
			Cityscapes		  &74.32  	&79.90    &80.87  	&\textbf{81.33}\\
			\bottomrule
		\end{tabular}
	\end{center}\label{tab:backbone}
\end{table}

\subsubsection{Study on the contribution of the auxiliary loss}

This section evaluates the effect of introduced auxiliary loss $\mathcal{L}_i$, which is helpful to optimize the whole training process, and has no interference with learning the master branch loss $\mathcal{L}_f$. By adjusting hyper-parameter $\lambda$ in range [0, 0.9] in steps 0.1, we carry on a set of experiments using ResNet-101 backbone, together with the master branch of \emph{BiCANet} for optimization. Note $\lambda = 0$ indicates the entire training process is dominant by master branch loss $\mathcal{L}_f$. The results is shown in Table \ref{tab:Loss}. It is observed that, when $\lambda = 0.1$, adding auxiliary loss $\mathcal{L}_i$ yields the best performance of 79.9\% mIoU on Cityscapes dataset. Along with the increase of $\lambda$, the performance drops significantly. Following \cite{zhao2017pyramid}, we believe deeper networks will benefit more from the new augmented auxiliary loss.

\begin{table}[!t]
	\tabcolsep 1.1mm \caption{Experiments on the contribution of the auxiliary loss to our \emph{BiCANet}. The bold number indicates the best performance in terms of mIoU.}
	\begin{center}
		\begin{tabular}{c|cccccccccc}
			\toprule
			$\lambda$ 	&0 	&0.1  &0.2	&0.3 &0.4 	&0.5  &0.6	&0.7	&0.8	&0.9\\
			\midrule
			Cityscapes		  		&74.3  	&\textbf{79.9}    &79.4  	&78.7	&78.1  	&77.7    &77.4  	&76.9	&76.2  	&75.3\\
			\bottomrule
		\end{tabular}
	\end{center}\label{tab:Loss}
\end{table}

\subsubsection{Ablation study for augmented training data}

Deep neural networks are data-hungry models, thus whether training data are enough or not plays an essential role for the performance. This section measures this effect by considering the augmented training data. Table \ref{tab:Data} exhibits the ablation results on Cityscapes dataset using different augmented settings, such as random scaling (RS), aspect ratio (AR), and image flipping (IF), as mentioned before. The experimental results show that, using all augmentation approaches, \emph{BiCANet} achieves the best performance, yielding 9.23\% mIoU improvement over the pre-trained model on ImageNet. It is also shown that each of these augmentation methods consistently improves the performance, improving segmentation results by 4.57\%, 3.37\%, and 1.29\% of mIoU, respectively.

\begin{table}[!t]
	\tabcolsep 5.5mm \caption{Ablation experiments using different data augmented methods on Cityscapes \emph{validation} set. The bold number indicates the best performance in terms of mIoU.}
	\begin{center}
		\begin{tabular}{c|ccc|c}
			\toprule
			Backbone  &RS &AR &IF &mIoU\\
			\midrule
			ResNet-101 &~		 	&~			&~		 	&70.67   \\	
			ResNet-101 &\checkmark 	&~			&~       	&75.24  \\
			ResNet-101 &\checkmark	&\checkmark	&~	  	 	&78.61  \\			
			ResNet-101 &\checkmark  &\checkmark &\checkmark &\textbf{79.90}  \\
			\bottomrule
		\end{tabular}
	\end{center}\label{tab:Data}
\end{table}

\section{Conclusion Remarks and Future Work}\label{sec:Conclusion}

In this paper, we have presented a novel network architecture, \emph{BiCANet}, for the task of semantic segmentation. \emph{BiCANet} aggregates context information form categorical perspective. At beginning, the subnetwork of CCPB learns a powerful category-based projection that combines features with different receptive fields. Thereafter, the subnetwork of BCIB enhances the context interaction within different intermediate convolution layers. Finally, the subnetwork of MCFB aggregates multi-scale contextual cues from local surroundings to long-ranged dependencies, even to the global context. We have evaluated the proposed \emph{BiCANet} on PASCAL VOC 2012, Cityscapes, and ADE20K datasets. The exhaustive experimental results show the superior performance of \emph{BiCANet} over recent state-of-the-art networks, and demonstrate that our approach can produce more consistent segmentation predictions with accurately delineated object shapes and boundaries. Moreover, it does not require any CRF post-processing. 

In the future, there are two aspects that we are interested to improve upon. The first one is to design a lightweight version of the proposed network, which satisfies the real-time requirement of real-world applications. On the other hand, in spite of achieving promising results for semantic segmentation, we believe that our method can be easily transferred to any existing network architectures that are used for other visual tasks, such as object detection \cite{redmon2017yolo9000}, saliency detection \cite{cheng2011global}, and depth estimation \cite{chen2019to}.

\appendices



\ifCLASSOPTIONcaptionsoff
  \newpage
\fi



%

\bibliographystyle{IEEEtran}
\bibliography{refs}

%








\end{document}